%% file: samplepaper.tex
\documentclass[]{article}
\usepackage[T1]{fontenc}
\usepackage{graphicx}
\input{header.tex}

\usepackage[a4paper, total={6in, 8in}]{geometry}
\usepackage{etoolbox}
\makeatletter
\providecommand{\institute}[1]{
  \apptocmd{\@author}{\end{tabular}
    \par
    \begin{tabular}[t]{c}
    #1}{}{}
}
\makeatother

\title{Combining self-labeling and demand based active learning for 
non-stationary data streams}
\author{Valerie Vaquet (\texttt{vvaquet@techfak.uni-bielefeld.de}),\\
Fabian Hinder, Johannes Brinkrolf, and
Barbara Hammer}
\institute{Machine Learning Group, Bielefeld University, 33501 Bielefeld, Germany}
\date{February 2023}
\begin{document}
%
%
\maketitle              
\begin{abstract}
Learning from non-stationary data streams is a research direction that gains increasing interest as more data in form of streams becomes available, for example from social media, smartphones, or industrial process monitoring.
Most approaches assume that the ground truth of the samples becomes available (possibly with some delay) and perform supervised online learning in the test-then-train scheme. While this assumption might be valid in some scenarios, it does not apply for all settings. In this work, we focus on scarcely labeled data streams and explore the potential of self-labeling in gradually drifting data streams.
We formalize this setup and propose a novel online $k$-nn classifier that combines self-labeling and demand based active learning. 
\end{abstract}
\section{Introduction}
Machine learning models are applied in a great variety of tasks ranging from image recognition in medicine and industrial settings to machine translation. They are reshaping our way of working and living -- for example by automating quality control, improving the individualization of medical support and making autonomous driving possible. As machine learning is applied in an ever changing world, a key requirement for machine learning is their adaptability to changes in the environment \cite{DBLP:journals/ijon/LosingHW18}. Such changes, also referred to as concept drift, can have different origins. They can be induced by changes of the observed environment or behavior -- for example by different light conditions over the day or changes in user behavior due to societal trends -- or by the technologies used to observe the environment -- for example by failing or degrading sensors.

Considerable research has been conducted on learning from non-stationary data streams~\cite{gama_survey_2014,DBLP:journals/sigkdd/GomesRBBG19}. Most approaches target the supervised online learning set-up which means that after the prediction of a sample, the ground truth becomes available (possibly with some delay) and can be used for the model update. In many real-world applications such as quality control, the ground truth information is not generated automatically. Rather, it requires time consuming measurements and/or expert knowledge. This limits the applicability of these technologies in real-world applications.

There exist contributions on learning from scarcely labeled data streams which can be categorized into two main directions: In semi-supervised online learning, the assumption is that some ground truth information becomes available. Usually, approaches combine the labeled and unlabeled information for updating the online learner~\cite{masud_facing_2012,dyer_compose_2014,DBLP:journals/prl/ZhuSH14}.
Alternatively, authors rely on active learning (\AL), which is an established approach in batch learning~\cite{settles_active_2009}. The basic idea is that the model can request ground truth information up to a certain budget to update the model. \AL strategies often select the most informative samples, i.e.\ those close to the decision boundary. While \AL is straightforward in batch learning,  online learning adds  additional requirements. For example, data which is not close to the decision boundary can be influenced by certain types of drift. Besides,  controlling the labeling budget throughout the data stream is crucial to keep an updated model even in the presence of concept drift~\cite{zliobaite_active_2014}. 
Some works consider the combination of semi-supervised and \AL~\cite{goldberg_oasis_2011,SSL_deep}.
Many \AL strategies discard unlabeled samples, albeit this might induce a loss of valuable information~\cite{DBLP:books/crc/aggarwal14/AggarwalKGHY14}. According to Fahy et al.~\cite{fahy_scarcity_2023} so-called self-labeling (\SL) is one possible future research direction for using the unlabeled samples. \SL strategies based on
majority clustering, label prediction, or more deep generative modeling have been proposed in the context of online learning~\cite{MOA,LI2019383}.
Korycki et al~\cite{korycki_combining_2018} demonstrate that a \SL step can enhance the performance in online \AL for some data sets. They rely on a fixed labeling budget for the \AL step and add an option for \SL if a sample is not chosen for \AL.

As finding a suitable budget is very data specific, we instead propose to rely on a combination of \SL and a demand driven \AL procedure. Our presented novel methodology bases the demand for labeling on the properties of the drift. Besides, it incorporates \SL together with a particularly efficient confidence estimation and labeling based on a distance and time-dependent weighting scheme. The suitability of this modeling is motivated by different types of drift and evaluated in comparison to the work by Korycki et al.~\cite{korycki_combining_2018} and several baselines.

The paper is organized as follows. First, we define the considered set-up and recall the definition of concept drift (see Section~\ref{sec:setup}). Then, we motivate using \SL for learning from non-stationary data streams from a theoretical perspective (Section~\ref{sec:theory}). After proposing a categorization of drift behaviors and the induced challenges for self labeling (Section~\ref{sec:drift-types}), we describe the proposed online learning framework (Section~\ref{sec:method}). Finally, we describe our experimental evaluation, present the results (Section~\ref{sec:exp}), and conclude the work (Section~\ref{sec:concl}).

\section{Set-Up\label{sec:setup}}
While in batch learning all data is available in the beginning,
in online learning the data is arriving in a streaming fashion: 
\begin{equation}
    S={(x_t, y_t)}_{t=1}^T,\quad x_t\in X,\quad y_t\in Y,\quad t=1,\dots, T 
\end{equation}
with input set $X$, label set $Y$, and arrival time of the data point $t$. The learning algorithm incrementally infers a model $h_t$ which approximates the unknown data distribution $P_t(X,Y)$ at time step $t$.
%
Non-stationarity refers to the fact that the data distribution might change over time. More formally, if 
$\exists t_0, t_1: P_{t_0}(X,Y)\neq P_{t_1}(X,Y)$
we say that concept drift is present in the data stream. This drift can be categorized into real and virtual drift. While for virtual drift only the distribution of the labels $P(Y)$ or the conditional probabilities $P(X|Y)$ change without changing the decision boundary, in real drift a change in the posterior $P(Y|X)$ and thus in the decision boundary is observed. Additionally, one distinguishes drift behavior into sudden or abrupt drift where the underlying distributions change significantly  from one time step to the next, incremental drift where the distribution is shifting continuously, and gradual drift where samples are generated by two distributions in an alternating fashion for the time of the drift \cite{gama_survey_2014}.

Within incremental learning, the ground truth label for updating the model at each time step might not be available in some scenarios. We
will consider the following scenario:
We assume an initial training phase where labels are available, and a subsequent life-long adaptation phase after the model has been deployed. Here, labels are available  after querying only.  
Hence, we assume a split of the data stream into two parts:
The first part of the stream $S_{\text{train}}$ contains labeled samples for both testing and updating in the interleaved fashion. In contrast, the evaluation part $S_{\text{test}}$ only contains ground truth for testing purposes. The training has access to the input samples $x_t$; 
Label information is available after querying only if necessary. 
This set-up mirrors popular assumptions for semi-supervised online learning, which often assume that initially a labeled batch of data is available \cite{dyer_compose_2014}.  

There exist evaluation schemes that are particularly suited for streaming data:
For the interleaved train test error (ITTE) the model $h_t$ is first evaluated and then updated for each time step $t$. This way, one can evaluate the overall performance until time $T$:
\begin{equation}
   E_S(h)=\frac{1}{T}\sum_{t=1}^T \ell(h_t(x_t), y_t),
\end{equation}
where $\ell$ refers to the loss function of the model output at time step $t$, e.g.\ the zero-one loss for classification tasks. The accuracy for model $h$ on the entire data stream $S$ can then be obtained by $A_S(h) = 1-E_S(h)$. 

\section{Motivation of self-labeling procedure\label{sec:theory}}

An upper bound on the model error~\cite{mohri2012new} in the context of concept drift can be given in terms of a comparison of the generalization error (risk) of the model $h$ and a (model specific) distributional discrepancy:
\newcommand{\apply}{\text{te}}
\newcommand{\train}{\text{tr}}
\begin{align*}
    \underbrace{E_\apply(h)}_{\text{application time error}} \leq\:\: \underbrace{E_\train(h)}_{\text{train time error}} +\quad \underbrace{\sup_{h' \in \mathcal{H}} \left| E_\apply(h') - E_\train(h')\right|}_{\text{distributional discrepancy}},
\end{align*}
where the supremum is taken over all considered models  $\mathcal{H}$, respectively. 
Here the risk might be the expected loss of a fixed model or the interleaved train-test error of an incremental learner, as introduced before.
For a static model, the discrepancy usually grows comparably fast as the distributions diverge over time. 
Hence we focus on adaptive methods, i.e.\ 
we refer to incremental learning schemes rather than static models. Still, learning such adaptive schemes is 
difficult or ill-posed, respectively, if important information such as the label information is not provided or restricted. Therefore, we will rely on specific assumptions regarding the underlying distribution as well as the occurring drift.

As a motivation, consider the following (parametric) example:
Assume that at every time point $t$ our input data $X$ is distributed according to a mixture model of uniform distributions on a unit ball with given center, assume that every cluster is associated with an unchanging label distribution $Q_i$, i.e., we have
\begin{align*}
    P_t(x,y) = \sum_{i = 1}^n \lambda_{i}(t) \mathcal{U}(x \mid \mu_{i}(t)) Q_i(y),
\end{align*}
where $\mathcal{U}(\: \cdot \mid \mu)$ denotes the uniform distribution on the unit ball around $\mu$ and $\lambda_i(t), \mu_i(t)$ are functions that represent the weight and mean of the $i$-th component at time $t$, respectively. In such a setup we need to identify the component $i$ that generated a sample $x$ and then obtain the label estimate $\hat{y} \sim Q_i$. 
The possibility of identifying the cluster relies on three assumptions: 
(i) all clusters move continuously, i.e. all $\mu_i$ are continuous, (ii) none of the clusters is vanishing, i.e.\ $\lambda_i > 0$, and (iii) the clusters do not merge, i.e.\ $\Vert \mu_i(t) - \mu_j(t) \Vert > 2$ for all $t$. Then we can trace back the path of the component through time to the point where we have knowledge about the label.




In practice, we usually do not observe continuous
parameter curves $\mu_i$ but we have to reconstruct them from data. To do so we rely on the fact that
observations form a manifold in data-space and time~\cite{hinder2020towards}. 
Hence prediction approaches can be traced back to the challenge of finding the closest labeled sample in terms of the \emph{intrinsic metric of the manifold}. 
A standard way of reconstructing a manifold (and its metric) from data can rely on the $k$-nn graph of (a subset of) the data points~\cite{doi:10.1126/science.290.5500.2319}. The distances in the manifold are then approximated based on distances in this graph. 
If enough data is provided, the connected components of this neighborhood graph correspond to mixture components of the underlying  distribution, characterizing labels $Q_i$. Thus, in this case, it is possible to uncover the label based on the connected components of the data.

We can extend this procedure  to more general data-time manifolds which do not strictly obey the assumptions. In this case, we can no longer assure that we identify the correct label if the clusters overlap or collide.
In such cases, and only in such cases, additional labeling is required.

In the following, we will rely on $k$-nn-based modeling and \SL to approximate the intrinsic metric, see Section~\ref{sec:method}.  We enhance this approach by active labeling in case of collisions, which we characterize in the next section.

\vspace*{-1em}
\section{A taxonomy of possible drift behavior\label{sec:drift-types}}
\vspace*{-0.5em}

\begin{figure}[t]
     \centering
     \begin{subfigure}[b]{0.43\textwidth}
         \centering
         \includegraphics[width=\textwidth]{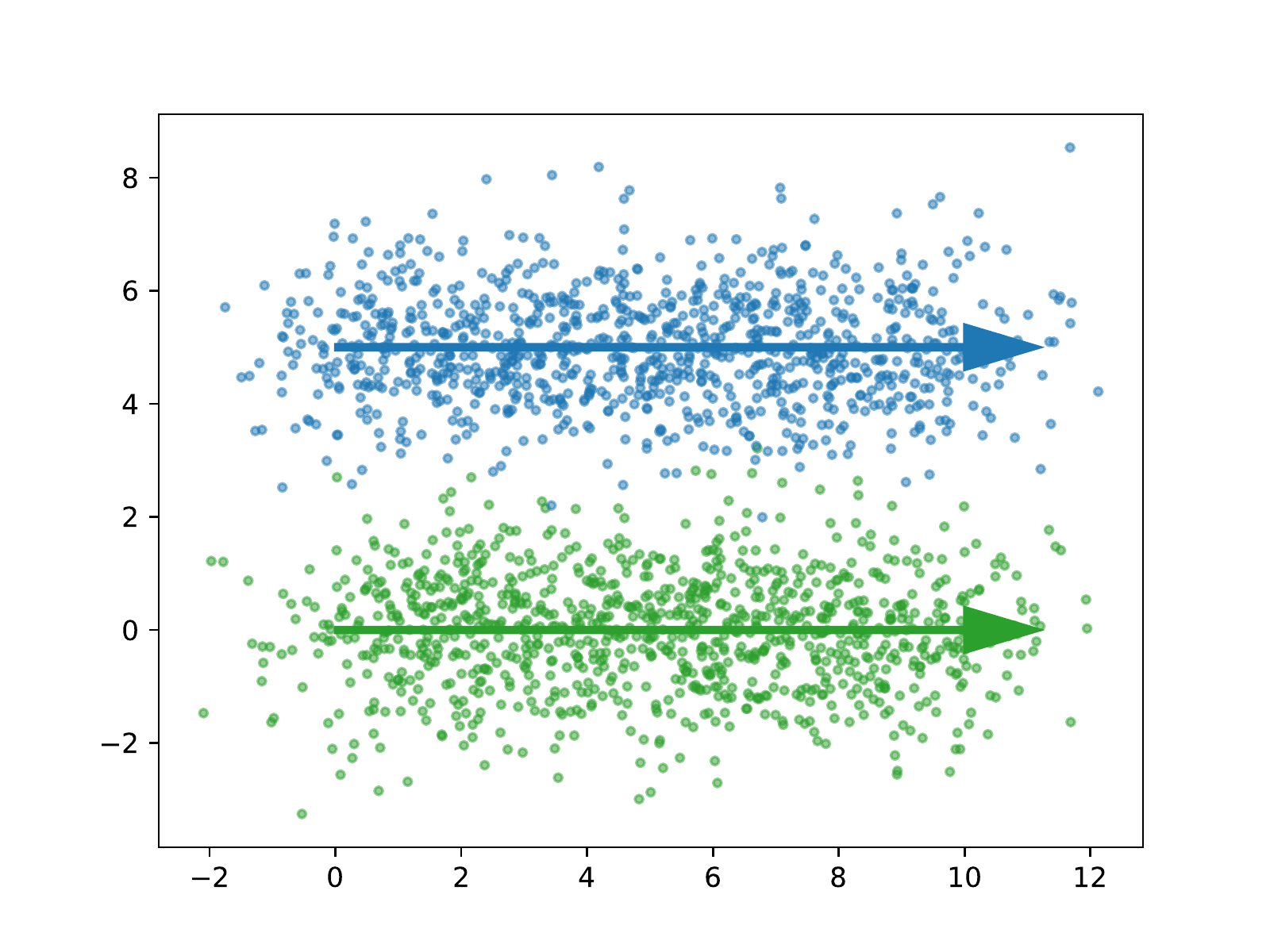}
         \caption{no overlap}
         \label{fig:easy}
     \end{subfigure}
     \begin{subfigure}[b]{0.43\textwidth}
         \centering
         \includegraphics[width=\textwidth]{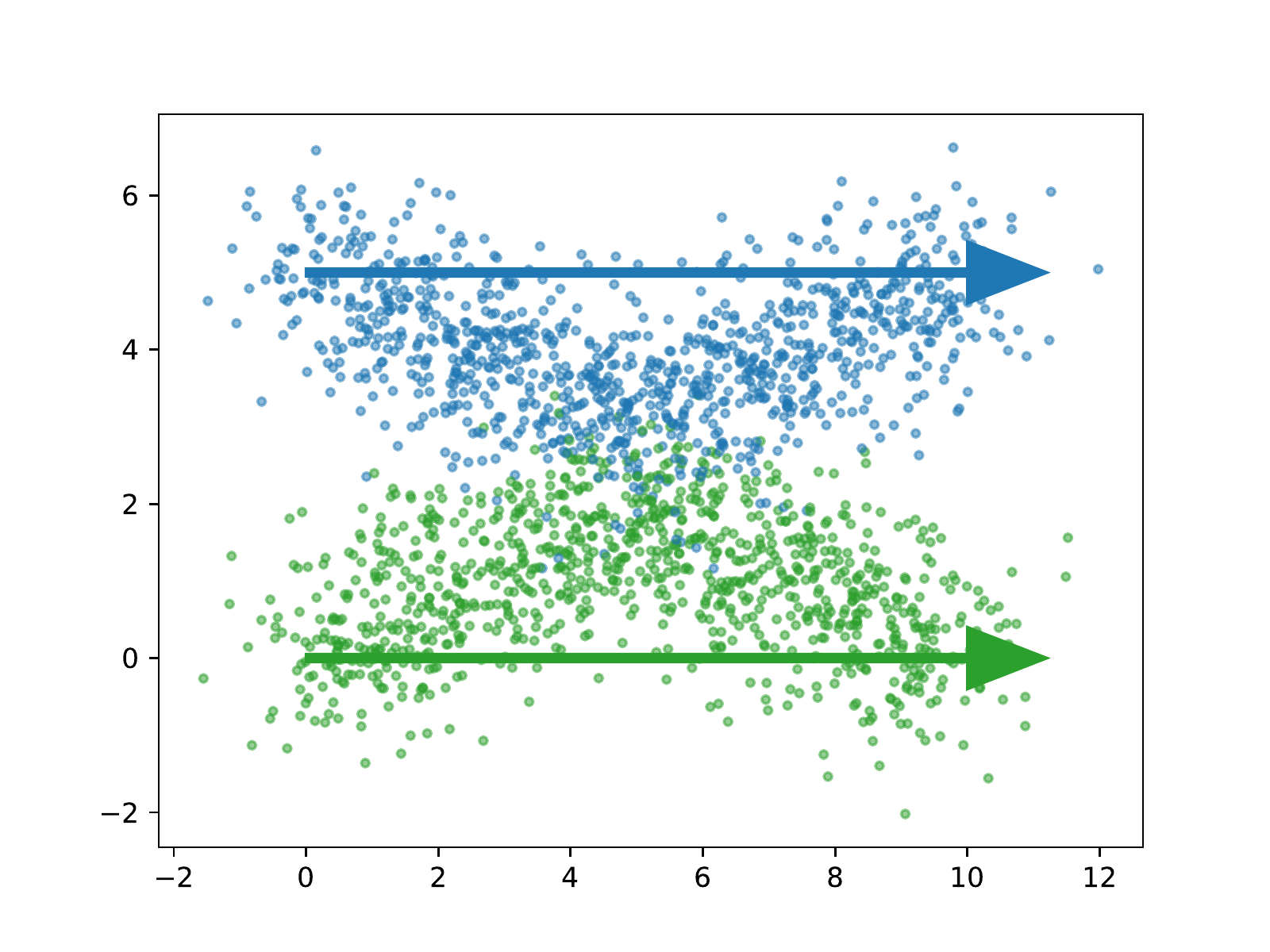}
         \caption{partly overlapping}
         \label{fig:overlapping}
     \end{subfigure}
     
     \begin{subfigure}[b]{0.43\textwidth}
         \centering
         \includegraphics[width=\textwidth]{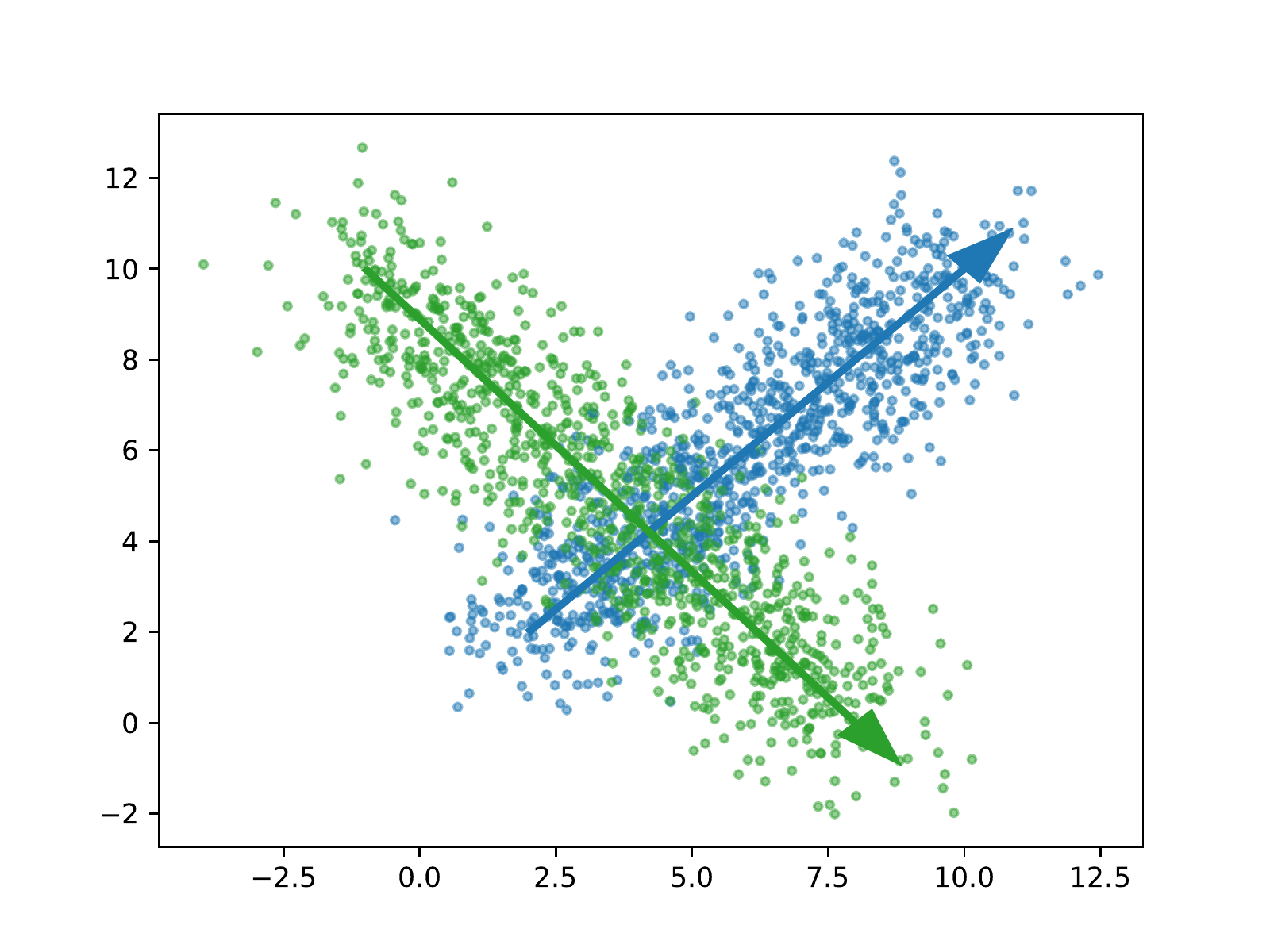}
         \caption{crossing at different times}
         \label{fig:crossing}
     \end{subfigure}
     \begin{subfigure}[b]{0.43\textwidth}
         \centering
         \includegraphics[width=\textwidth]{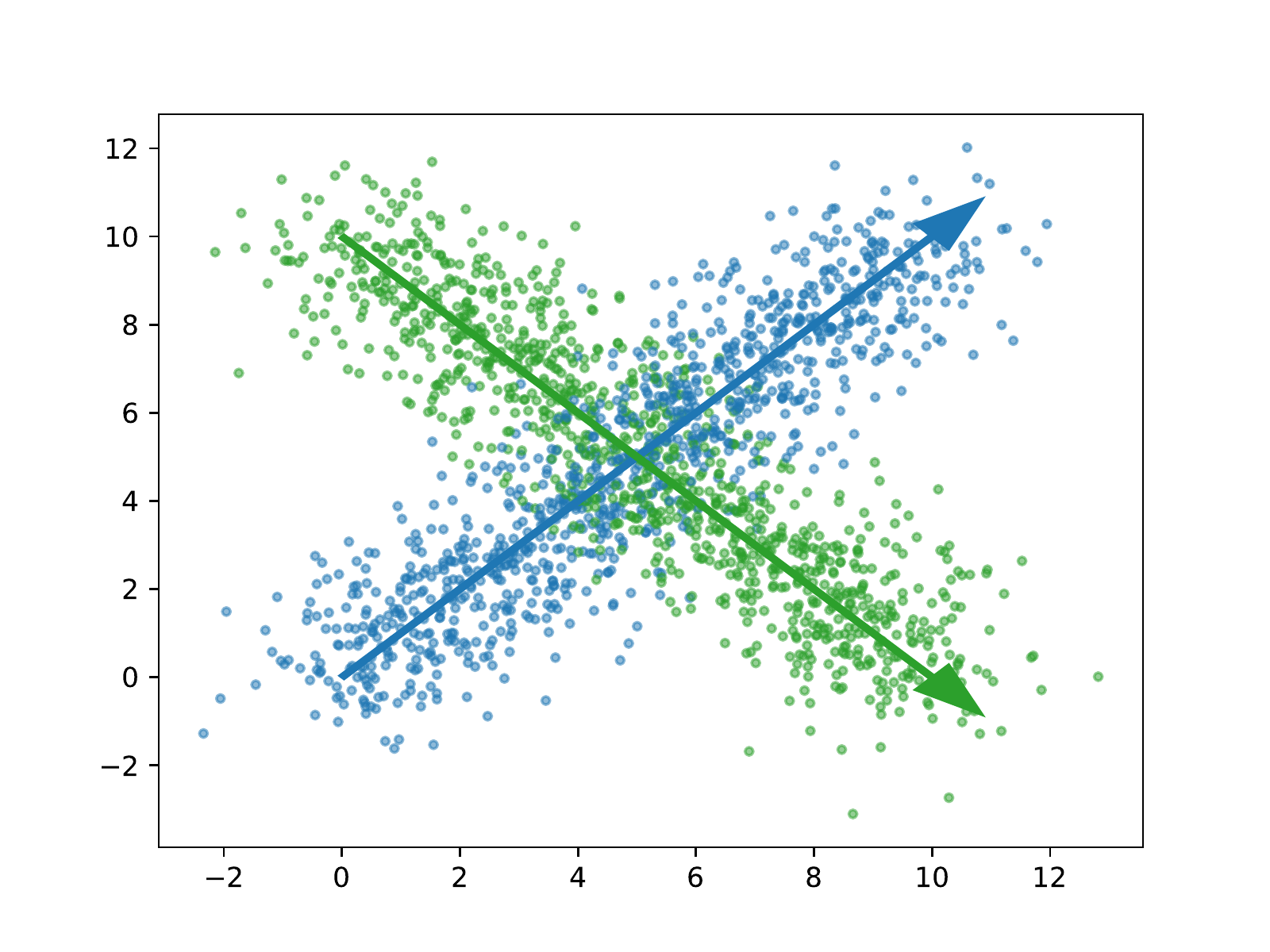}
         \caption{crossing at the same time}
         \label{fig:crossing-at-same-time}
     \end{subfigure}
        \caption{Visualization of the considered drift behaviors}
        \label{fig:data-streams}
 \vspace*{-2em}
\end{figure}

In this section, we categorize different drift behaviors w.r.t. relevant attributes regarding a \SL $k$-nn based online learner as summarized in Section~\ref{sec:method}.
%
%
%
%
Restricting to gradual drift, we decompose its shape into four main scenarios:
\begin{enumerate}
    \item the underlying distributions are moving without any overlap (Figure~\ref{fig:easy}),\label{scenario-easy} 
    \item the underlying distributions are partly overlapping at certain times, but at each time step there are clearly separable regions (Figure~\ref{fig:overlapping}),\label{scenario-ovelapping}
    \item the underlying distributions both occupy some regions of the space at different points in time  (Figure~\ref{fig:crossing}),\label{scenario-crossing}
    \item the underlying distributions are overlapping in space at the same time point (Figure~\ref{fig:crossing-at-same-time}).\label{scenario-crossing-same-time}
\end{enumerate}
Based on our considerations in Section~\ref{sec:theory}, we expect that the first case works well relying on \SL only as the connected components in the neighborhood graph decompose according to the initial label.  While Scenario~\ref{scenario-ovelapping} might cause some challenges, an unsupervised adaptation of models and good classification results are possible. 
In contrast to the first scenarios, the latter two cannot rely on \SL alone if the two components  of the manifold overlap corresponding to the labeling. The third scenario can resolve spatial conflicts by classification based on adapted window sizes. The last scenario contains an overlapping area in which the label distributions strictly merge. Thus, it is not clear which prediction is correct. Here, \AL can help to identify the correct labels for the individual components after the overlap has happened.

In the following section, we will first introduce an approach relying on \SL and \AL on demand, which accounts for the discussed drift behavior. Then, we will  demonstrate its performance in a number of experiments.

\section{Self-labeling framework\label{sec:method}}
The basic idea of \SL is to consider the model prediction as the ground truth label for model update provided the prediction has a reasonably high certainty \cite{DBLP:journals/prl/ZhuSH14}. 
As already discussed in the last section, we expect that additional ground truth information is necessary for more challenging drift scenarios. Hence, we complement \SL with \AL steps when required.

In the following, we rely on an online $k$-nn classifier since this seems particularly suited for incremental learning \cite{DBLP:journals/ijon/LosingHW18}. While there are popular versions working with multiple memories storing different concepts~\cite{DBLP:journals/kais/LosingHW18} or adapting the size of a sliding window when drift is detected~\cite{DBLP:conf/sdm/BifetG07}, applying these in the considered setting of scarcely labeled data is not possible as they are relying on ground truth information. Therefore, we consider a sliding window $k$-nn model with a fixed window size. This approach naturally leads to extensions for \SL in incrementally shifting data streams.
As a \emph{decision rule}, we propose an adaption of the commonly used weighted $k$-nn decision rule:
\begin{align}
    h(x)=\hat{y}= \argmax_{c\in\{1,\,\dots,\,C\}} \sum_{(x_i, y_i)\in N_k(x)}\gtweight\ageweight\delta_{y_i, c} \frac{1}{\text{dist}(x_i, x)}.\label{eq:pred}
\end{align}
Here, $\delta_{l,m}$ is the Kronecker delta, $\text{dist}(x_i, x)$ is the distance between $x_i$ and $x$, and $\gtweight$ and $\ageweight$ are sample specific weights which account for the fact that more recent samples have a higher relevance. Setting them to one leads to the vanilla decision rule over the considered points. They will be introduced in more detail at the end of this section. 
The corresponding \emph{certainty} of the classification is given as
\begin{align}
    c(x, \hat{y})= \frac{1}{\gtnorm \agenorm\sum_{x_i\in N_k(x)} \frac{1}{\text{dist}(x_i,x)}}\sum_{(x_i, y_i)\in N_k(x)}\gtweight\ageweight\delta_{y_i, \hat{y}}\frac{1}{\text{dist}(x_i, x)},
   \label{eq:certainty}
\end{align}
where $\gtnorm$ and $\agenorm$ are normalizing factors.

\begin{algorithm}[t]
\caption{Hybrid \SL / active stream learning}\label{algo:alg}
\textbf{Input:} streams $S_\text{train}$, $S_\text{test}$ model $h$, $\alwindow$, $\thresSL$, $\thresAL$
\begin{algorithmic}[1]
\State init list certainties
\For{$(x_t, y_t)\in S_\text{train}$}
\State call Algorithm~\ref{algo:train}
\EndFor
\State $\thresSL = $ 25th quartile of certainties
\For{$x_t\in S_\text{test}$}
\State call Algorithm~\ref{algo:test}
\EndFor
\end{algorithmic}
\end{algorithm}

\begin{algorithm}[t]
\caption{Learning Step in $S_\text{train}$}\label{algo:train}
\textbf{Input:} sample $(x,y)$, model $h$, list certainties \\
\textbf{Output:} model $h$, certainties
\begin{algorithmic}[1]
\State $\hat{y}=h(x)$ according to Eq.~\ref{eq:pred}
\State $c=c(x,\hat{y})$ according to Eq.~\ref{eq:certainty}
\State update $h$ with $(x,y)$
\If{$y=\hat{y}$}
\State append $c$ to certainties 
\EndIf
\end{algorithmic}
\end{algorithm}

Next to the certainty scores, a mechanism to decide whether a prediction is confident enough for including the sample and predicted label in the  model update is required. 
We can rely on  characteristics  acquired from the training stream $S_{\text{train}}$ for setting this \emph{threshold} $\thresSL$. More precisely during this phase, one can collect the certainties of correct predictions and rely on statistics, e.g.\ the mean or a lower quartile, as an estimate of how confident a model is when predicting correctly (see Algorithm~\ref{algo:alg}).

As already described in Section~\ref{sec:setup}, during $S_\text{train}$ the classifier will be updated with $(x_t, y_t)$ for each time step $t$ (see Algorithm~\ref{algo:train}). In contrast during $S_\text{test}$ an \emph{update} with $(x_t, \hat{y}_t)$ will only take place if $c(x,\hat{y})>\thresSL$ (see Algorithm~\ref{algo:test}).

As we are considering different types of non-stationary data streams, we might arrive at a point where \SL is not sufficient to account for drifts of the underlying data distribution. This is reflected by a decreasing certainty of the model's predictions. 
If 
$\frac{1}{\detectwindow}\sum_{t-\detectwindow}^tc(x_t,\hat{y}_t)<\thresAL$, i.e. the mean certainty of a window of the last $\detectwindow$ samples drops under a pre-defined threshold $\thresAL$, we interrupt the \SL procedure and obtain ground truth by \emph{\AL} (compare lines 8-10 in Algorithm~\ref{algo:test}).

We do not apply any labeling budget, since we can limit the possibility to initiate \AL by choosing a small window of size $\alwindow$ during which acquiring ground truth is possible. As we want to analyze the potential of requesting the labels while keeping the number of requested labels small, we experiment with two configurations: In the first one, the model can perform \AL within a fixed time window of $\alwindow$ samples if the certainty is too small to perform self-learning. In contrast, in the second, we make the ground truth information of the next $\alwindow$ samples available to the model.

\begin{algorithm}[t]
\caption{Learning Step in $S_\text{test}$}\label{algo:test}
\textbf{Input:} sample $x$, label $y$ if requested, model $h$, $\alwindow$, $\currentalwindow$, $\thresSL$, $\thresAL$ \\
\textbf{Output:} model $h$
\begin{algorithmic}[1]
\State $\hat{y}=h(x)$ according to Eq.~\ref{eq:pred}
\If{$c(x,\hat{y})>\thresSL$}
\State update $h$ with $(x,\hat{y})$ 
\ElsIf{$\currentalwindow >0$}
\State \AL: request $y$, update $h$ with $(x,y)$ 
\EndIf
\State decrease $\currentalwindow$ by 1 
\If{$\frac{1}{\detectwindow}\sum_{t-\detectwindow}^t c(x_t,\hat{y}_t)<\thresAL\And \currentalwindow\leq0$}
\State $\currentalwindow=\alwindow$
\EndIf
\end{algorithmic}
\end{algorithm}

In particular, for more challenging drift scenarios, weighting samples contained in the sliding window is beneficial. Especially when the distributions are crossing with a small time delay (compare Scenario~\ref{scenario-crossing}), more recent samples are more informative than older ones to make a reliable decision. Therefore, we propose to \emph{weight} sample $x_i$ with weight $\ageweight$, which is 1 if $x_i$ is the most recent sample and the weight is linearly decaying for older samples in the sliding window. The normalizing factor $\agenorm$ is the mean of the $k$ maximal weights $\ageweight$.

Working with \SL and making \AL possible if the model does not provide predictions with a high enough certainty, it might be advantageous if follow-up predictions consider whether a sample is added to the $k$-nn window with a ground truth or a prediction based label. We propose to weight ground truth samples with a higher weight $\gtweight$ than others in the prediction of new samples and the estimation of the according certainty score. Again, $\gtnorm$ is the mean of the $k$ maximal weights $\gtweight$.

\section{Experiments\label{sec:exp}}


We experimentally evaluate our proposed method on the scenarios described in Section~\ref{sec:drift-types} and on a real-world data set.
For the experiments on artificial data, we generate 100 sample streams containing 2000 instances for each scenario described in Section~\ref{sec:drift-types} and visualized in Figure~\ref{fig:data-streams}. The data streams are generated by drawing samples from a normal distribution with an incrementally shifting mean which is visualized by the arrows.
Additionally, we consider a real-world spam data set with underlying gradual drift that contains  9,324 samples~\cite{DBLP:journals/jiis/KatakisTBBV09}. Since the data set is very high-dimensional (39,916 features), we first apply a dimensionality reduction by PCA to obtain a representation of 10 features. The PCA is fitted on $S_\text{train}$.

\begin{table}[h]
    \centering
 	\addtolength{\tabcolsep}{2pt}
 \tiny
     \tabfaultdet{res/results_a_latexvodoo.csv}{}
     \caption{Scenario~\ref{scenario-easy}\label{tab:res1}}
 \end{table}
 \begin{table}[h]
 \vspace*{-2em}
    \centering
 	\addtolength{\tabcolsep}{2pt}
 \tiny
    \tabfaultdet{res/results_b_latexvodoo.csv}{}
\caption{Scenario~\ref{scenario-ovelapping}\label{tab:res2}}
 \vspace*{-3.5em}
 \end{table}
 \begin{table}[h]
    \centering
 	\addtolength{\tabcolsep}{2pt}
 \tiny
 \tabfaultdet{res/results_c_latexvodoo.csv}{}
 \caption{Scenario~\ref{scenario-crossing}\label{tab:res3}}
 \end{table}
 \begin{table}[h]
 \vspace*{-2em}
    \centering
 	\addtolength{\tabcolsep}{2pt}
 \tiny
 \tabfaultdet{res/results_d_latexvodoo.csv}{}
  \caption{Scenario~\ref{scenario-crossing-same-time}\label{tab:res4}}
 \vspace*{-3.5em}
 \end{table}
 \begin{table}[h]
    \centering
 	\addtolength{\tabcolsep}{2pt}
 \tiny
 \tabfaultdet{res/results_spam_latexvodoo.csv}{}
 \caption{Real world: spam\label{tab:realworld}}
 \vspace*{-3.5em}
\end{table}

We compare the results for the versions of our proposed approach relying on \SL with \AL (SL+\AL) and \SL with a batch update (SL+batch) respectively to the combination of \AL and \SL (AL+\SL) with labeling budgets of $1\%$ and $5\%$ as proposed by \cite{korycki_combining_2018}. We additionally report the results for learning from the fully supervised stream (supervised), for vanilla \SL, and \AL with fixed labeling budgets of $1\%$ and $5\%$\footnote{We set the hyperparameters as follows: $k=5$; window size of the $k$-nn algorithm $\windowsize=100$; $\thresSL=\,$25th quartile of collected certainties; $\detectwindow=\frac{1}{5}\windowsize$; $\thresAL=0.9\thresSL$;$\alwindow=\detectwindow$. The aging weight for the most recent sample is set to 1, the one of the oldest to 0.9, in between the weights are linearly interpolated.}. The results are summarized in Table~\ref{tab:res1}-\ref{tab:realworld}. We report the mean and standard deviation of the accuracy and the mean of the requested samples over 100 experimental runs. 
We analyze the performance of \SL on each data set individually and discuss the effect of the proposed weightings. 
%

Considering Scenario 1 (see Table~\ref{tab:res1}) -- the case where the data distributions are co-existing without any overlap -- we obtain an accuracy of 0.99 for most settings which is the performance that can be obtained by the fully supervised learner. As to be expected, in this setting it is not necessary to request labeled information by \AL to update the model in a reasonable way. The only models performing slightly weaker are the purely \AL based approaches. This observation can be explained by the fact that these methods only use 1\% resp. 5\% of the data stream and discard all unlabeled information.


In the experiment with a partial overlap (compare Table~\ref{tab:res2}) we find that \SL is sufficient. Only in some rare cases, \AL is initiated resulting in very few requested samples for the version not triggering updating with a batch of labels. The reported classification prescription is comparable to the ground truth baseline and exceeds the one observed for \AL with a small labeling budget of 1\%. In this scenario, we further can confirm that adding \SL to \AL on a fixed budget improves the results for a small labeling budget from about 71\% to 75\%.


The results of our experiment for Scenario 3 are presented in Table~\ref{tab:res3}. As discussed before in this case the challenge is that we still have samples of the blue class in our $k$-nn memory which might confuse the classifier which is relying on scarce ground truth obtained by \AL and on its own predictions (compare Figure~\ref{fig:crossing-at-same-time}). Since the ground truth baseline can rely on label information for all samples it does not get confused by this scenario and obtains an accuracy of about 96\%. In contrast, we report an accuracy of about 71\% for the \SL framework. This can be increased to 75\% by enforcing an update with ground truth information as soon as the \SL scheme becomes too uncertain (SL+batch). This result is comparable to the AL scheme with a labeling budget of 1\%, while on average only using a few more ground truth samples.
Assigning more relevance to recent samples by introducing $\ageweight$ improves the results for self-AL and self-batch as introducing the weight makes the predictions for the outdated class less certain. 
In this scenario, one can also nicely see that here the combination of \AL and \SL as proposed by \cite{korycki_combining_2018} is not advantageous as it faces the same challenges as our approaches. Therefore, introducing the weight for sample age is also increasing the performance for the combination of AL and \SL in this case.


Finally, considering the most challenging scenario, where the distributions are fully overlapping at the same time (see Table~\ref{tab:res4}), we report considerably lower accuracies for all methods in comparison to the fully labeled setting. Again, the combination of \AL and \SL is performing worse than relying solely on \AL. Here, we observe that adding the ground truth weight $\gtweight$ improves the classification prescription of our proposed framework, as it gives the classifier an orientation.

While the introduced weighting factors $\ageweight$ and $\gtweight$ clearly increase the performance in the more challenging drifting behaviors with colliding distributions, they have a negative impact on the performance of \SL in the simpler scenarios. This is probably due to the fact that more samples are rejected from training due to lower certainty scores -- especially in the setting with partly overlapping distributions in Scenario~\ref{scenario-ovelapping}. It is important to note, that we make the same observation for the \AL approaches on a budget.

On the real-world data stream, we report an accuracy of about 0.96 in the fully labeled setup. In contrast to a purely \SL approach for which we observe an accuracy of about 0.29, our methods as well as \AL perform at a reasonable level. While \AL with a budget of 5\% obtains an accuracy of 0.93, our method performs at a comparable level (0.86) to the \AL on a budget of 1\% (0.84) while using fewer labels. This effect becomes even more apparent when ground truth weighting is active.
Again, we observe that the combination of \AL and \SL as proposed by \cite{korycki_combining_2018} decreases the performance considerably for both investigated labeling budgets.


\section{Conclusion\label{sec:concl}}
In this work, we proposed a \SL $ k$-nn-based online learning model with demand based \AL for learning from incrementally drifting data streams. We experimentally confirmed that \SL can successfully deal with distributions that are not or only partly overlapping.
Additionally, scenarios with considerable overlap benefit from \AL on demand. 
Here, on demand labeling might be superior to a fixed budget to guarantee the availability of sufficient information.
Further work should explore the combination with unsupervised drift detection which could guide \AL in the presence of more abrupt drifts. Besides, adapting popular more sophisticated $k$-nn-based online learning schemes as \cite{DBLP:journals/kais/LosingHW18} to scarce label availability would increase their applicability.
\bibliographystyle{splncs04}
\bibliography{self-learning}
%




\end{document}

%% file: header.tex
\usepackage{xcolor}
\usepackage{amsfonts,amssymb,amsmath,array}

\usepackage{caption}
\usepackage{subcaption}
\usepackage{float}
\usepackage{graphicx}
\usepackage{algorithm}
\usepackage{algorithmicx}
\usepackage{algpseudocode}

\usepackage{csvsimple}
\usepackage{lscape}
\usepackage{todonotes} 
\usepackage{caption}

\DeclareMathOperator*{\argmax}{arg\,max}

\newcommand{\gtweight}{\omega_\text{GT}^i}
\newcommand{\gtnorm}{\omega_\text{GT}}
\newcommand{\ageweight}{\omega_\text{age}^i}
\newcommand{\agenorm}{\omega_\text{age}}
\newcommand{\windowsize}{\tau}
\newcommand{\detectwindow}{\tau_\text{stop}}
\newcommand{\currentalwindow}{\tau_\text{AL}'}
\newcommand{\alwindow}{\tau_\text{AL}}
\newcommand{\thresAL}{\theta_\text{AL}}
\newcommand{\thresSL}{\theta_\text{SL}}
\usepackage{xspace}
\newcommand{\AL}{AL\xspace}
\newcommand{\SL}{SL\xspace}

\usepackage{booktabs}
\usepackage{hyperref}
\usepackage{multirow}
\usepackage{multicol}
\usepackage{nicefrac}
\usepackage{varwidth}
\usepackage{pgfplotstable}
\pgfplotsset{compat=1.16}
\usepackage{tabularx}
\usepackage{adjustbox}
\makeatletter
\pgfplotsset{
    /pgfplots/table/omit header/.style={%
        /pgfplots/table/typeset cell/.append code={%
            \ifnum\c@pgfplotstable@rowindex=-1
                \pgfkeyslet{/pgfplots/table/@cell content}\pgfutil@empty%
            \fi
        }
    }
}
\makeatother

\usepackage[binary-units,group-separator={,}]{siunitx}

\newcolumntype{N}{S[table-format=1.3,round-mode=places,round-precision=3]}
\newcolumntype{O}{S[table-format=1.3]}
\newcolumntype{Q}{@{\,}c@{\,}S[table-format=1.3,round-mode=places,round-precision=3]@{}c@{\ }}

\newcolumntype{L}{S[table-format=4.1,round-mode=places,round-precision=1]}
\newcolumntype{K}{S[table-format=1.2]}
\newcolumntype{M}{@{\,}c@{\,}S[table-format=3.1,round-mode=places,round-precision=1]@{}c}

\newcommand{\tabfaultdet}[3][]{
    \ifthenelse{\equal{#3}{}}{}
    {{\centering\normalsize\textbf{#3}\\\medskip}}
    \pgfplotstabletypeset[
    col sep=comma,
    string type,
    columns/ALsetup/.style={string type,column type={@{\ }l@{\ }|}, column name={setup}},
    columns/now/.style={column type=Q,column name=},
    columns/nowsamples/.style={column type=M,column name=},
    columns/wgt/.style={column type=Q,column name=},
    columns/wgtsamples/.style={column type=M,column name=},
    columns/wage/.style={column type=Q,column name=},
    columns/wagesamples/.style={column type=M,column name=},
    columns/wgtage/.style={column type=Q,column name=},
    columns/wgtagesamples/.style={column type=M,column name=},
    columns/nowm/.style={column type=N,column name=\multicolumn{3}{c}{\textbf{Acc}}},
    columns/nowsamplesm/.style={column type=L,column name=\multicolumn{1}{c|}{\textbf{$\#$samp.}}},
    columns/wgtm/.style={column type=|N,column name=\multicolumn{3}{c}{\textbf{Acc}}},
    columns/wgtsamplesm/.style={column type=L,column name=\multicolumn{1}{c|}{\textbf{$\#$samp.}}},
    columns/wagem/.style={column type=|N,column name=\multicolumn{3}{c}{\textbf{Acc}}},
    columns/wagesamplesm/.style={column type=L,column name=\multicolumn{1}{c|}{\textbf{$\#$samp.}}},
    columns/wgtagem/.style={column type=|N,column name=\multicolumn{3}{c}{\textbf{Acc}}},
    columns/wgtagesamplesm/.style={column type=L,column name=\multicolumn{1}{c}{\textbf{$\#$samp.}}},
    columns/method/.style={column type=l,column name={\textbf{fault}}},
    columns={ALsetup, nowm, now, nowsamplesm, wgtm, wgt, wgtsamplesm, wagem, wage, wagesamplesm, wgtagem, wgtage, wgtagesamplesm},
    every head row/.style={after row=\midrule},
    every head row/.append style={before row={& \multicolumn{5}{c|}{\small no weights} & \multicolumn{5}{c|}{\small $w_{GT}$} & \multicolumn{5}{c|}{\small $w_{age}$} & \multicolumn{5}{c}{\small $w_{GT}$ and $w_{age}$}\\[0.5em]}},
    every row no 4/.append style={before row={[0.3em]}},
    every row no 6/.append style={before row={[0.3em]}},
    #1,
    create on use/now/.style={
      create col/assign/.code={%
        \ifnum\pdfstrcmp{\thisrow{nows}}{}=0
        \edef\entry{--&&}
        \else
        \edef\entry{$\pm$&\thisrow{nows}&}
        \fi
        \pgfkeyslet{/pgfplots/table/create col/next content}\entry
      }
    },
    create on use/nowsamples/.style={
      create col/assign/.code={%
        \ifnum\pdfstrcmp{\thisrow{nowsampless}}{}=0
        \edef\entry{--&&}
        \else
        \edef\entry{$\pm$&\thisrow{nowsampless}&}
        \fi
        \pgfkeyslet{/pgfplots/table/create col/next content}\entry
      }
    },
    create on use/wgt/.style={
      create col/assign/.code={%
        \ifnum\pdfstrcmp{\thisrow{wgts}}{}=0
        \edef\entry{--&&}
        \else
        \edef\entry{$\pm$&\thisrow{wgts}&}
        \fi
        \pgfkeyslet{/pgfplots/table/create col/next content}\entry
      }
    },
    create on use/wgtsamples/.style={
      create col/assign/.code={%
        \ifnum\pdfstrcmp{\thisrow{wgtsampless}}{}=0
        \edef\entry{--&&}
        \else
        \edef\entry{$\pm$&\thisrow{wgtsampless}&}
        \fi
        \pgfkeyslet{/pgfplots/table/create col/next content}\entry
      }
    },
    create on use/wage/.style={
      create col/assign/.code={%
        \ifnum\pdfstrcmp{\thisrow{wages}}{}=0
        \edef\entry{--&&}
        \else
        \edef\entry{$\pm$&\thisrow{wages}&}
        \fi
        \pgfkeyslet{/pgfplots/table/create col/next content}\entry
      }
    },
    create on use/wagesamples/.style={
      create col/assign/.code={%
        \ifnum\pdfstrcmp{\thisrow{wagesampless}}{}=0
        \edef\entry{--&&}
        \else
        \edef\entry{$\pm$&\thisrow{wagesampless}&}
        \fi
        \pgfkeyslet{/pgfplots/table/create col/next content}\entry
      }
    },
    create on use/wgtage/.style={
      create col/assign/.code={%
        \ifnum\pdfstrcmp{\thisrow{wgtages}}{}=0
        \edef\entry{--&&}
        \else
        \edef\entry{$\pm$&\thisrow{wgtages}&}
        \fi
        \pgfkeyslet{/pgfplots/table/create col/next content}\entry
      }
    },
    create on use/wgtagesamples/.style={
      create col/assign/.code={%
        \ifnum\pdfstrcmp{\thisrow{wgtagesampless}}{}=0
        \edef\entry{--&&}
        \else
        \edef\entry{$\pm$&\thisrow{wgtagesampless}&}
        \fi
        \pgfkeyslet{/pgfplots/table/create col/next content}\entry
      }
    },
    ]{#2}
}

\newcommand{\tabfaultdetstrategy}[3][]{
    \ifthenelse{\equal{#3}{}}{}
    {{\centering\normalsize\textbf{#3}\\\smallskip}}
    \pgfplotstabletypeset[
    outfile=tab-fault.tex,
    col sep=comma,
    string type,
    columns/pre/.style={column type=Q,column name=},
    columns/rec/.style={column type=Q,column name=},
    columns/f/.style={column type=Q,column name=},
    columns/meanpre/.style={column type=N,column name=\multicolumn{3}{c}{\textbf{precision}}},
    columns/meanrec/.style={column type=N,column name=\multicolumn{3}{c}{\textbf{recall}}},
    columns/meanf/.style={column type=N,column name=\multicolumn{3}{c}{\textbf{F1}}},
    columns/method/.style={column type=l,column name={\textbf{imputation strategy}}},
    columns={method, meanpre, pre, meanrec, rec, meanf, f},
    every head row/.style={after row=\midrule},
    #1,
    create on use/pre/.style={
      create col/assign/.code={%
        \ifnum\pdfstrcmp{\thisrow{stdpre}}{}=0
        \edef\entry{--&&}
        \else
        \edef\entry{$\pm$&\thisrow{stdpre}&}
        \fi
        \pgfkeyslet{/pgfplots/table/create col/next content}\entry
      }
    },
    create on use/rec/.style={
      create col/assign/.code={%
        \ifnum\pdfstrcmp{\thisrow{meanrec}}{}=0
        \edef\entry{--&&}
        \else
        \edef\entry{$\pm$&\thisrow{stdrec}&}
        \fi
        \pgfkeyslet{/pgfplots/table/create col/next content}\entry
      }
    },
    create on use/f/.style={
      create col/assign/.code={%
        \ifnum\pdfstrcmp{\thisrow{meanf}}{}=0
        \edef\entry{--&&}
        \else
        \edef\entry{$\pm$&\thisrow{stdf}&}
        \fi
        \pgfkeyslet{/pgfplots/table/create col/next content}\entry
      }
    },
    ]{#2}
}